\documentclass[letterpaper, 10 pt, conference]{ieeeconf}  

\IEEEoverridecommandlockouts                              

\usepackage{times}
\usepackage{amsmath}
\usepackage{calc}
\usepackage[T1]{fontenc}
\usepackage{amssymb}
\usepackage{amsmath}
\usepackage{amstext}
\usepackage{graphicx,color}
\usepackage{booktabs,tabularx}
\usepackage{multirow}
\usepackage{url}
\usepackage{tabularx}
\graphicspath{{images/}}
\usepackage{caption}
\usepackage{subcaption}
\usepackage{tabularx} 
\usepackage{wrapfig}
\usepackage{flushend}

\makeatletter
\let\NAT@parse\undefined
\makeatother

\newcolumntype{Y}{>{\centering\arraybackslash}X}

\newcommand{\ignore}[1]{}

\def\figref#1{Fig.~\ref{#1}}

\def\eqref#1{Eq.~(\ref{#1})}

\newcommand{\bz}{\mathbf{z}}
\newcommand{\bx}{\mathbf{x}}

\newcommand{\by}{\mathbf{y}}

\newcommand{\bd}{\mathbf{d}}
\newcommand{\bW}{\mathbf{W}}
\newcommand{\bg}{\mathbf{g}}

\usepackage{flushend}
\newcommand{\etal}{{\em et al.}}

\newcommand{\papertitle}{Multimodal Deep Learning for Robust RGB-D Object Recognition}

\title{\LARGE \bf
\papertitle
}

\author{Andreas Eitel\and Jost Tobias Springenberg\and Luciano Spinello\and Martin Riedmiller \and Wolfram Burgard
  \thanks{All authors are with the Department of Computer Science, University of Freiburg, Germany. This work was partially funded by the DFG
  under the priority programm ``Autonomous Learning'' (SPP 1527).
    \{eitel, springj, riedmiller, spinello, burgard\}@cs.uni-freiburg.de}
}

\begin{document}

\maketitle
\thispagestyle{empty}
\pagestyle{empty}

\begin{abstract}
Robust object recognition is a crucial ingredient of many, if not all, real-world robotics
applications. 
This paper leverages recent progress on Convolutional Neural Networks (CNNs) and proposes a novel RGB-D architecture for object recognition.
Our architecture is composed of two separate CNN processing streams -- one for each modality -- which are consecutively combined with a late fusion network.
We focus on learning with imperfect sensor data, a typical problem in real-world robotics tasks.
For accurate learning, we introduce a multi-stage training methodology and two crucial ingredients for handling depth data with CNNs. 
The first, an effective encoding of depth information for CNNs that enables learning without the need for large depth datasets. 
The second, a data augmentation scheme for robust learning with depth images by corrupting them with realistic noise patterns.
We present state-of-the-art results on the RGB-D object dataset~\cite{lai2011large} and show recognition in challenging RGB-D real-world noisy settings.

\end{abstract}

\IEEEpeerreviewmaketitle

\section{Introduction}

RGB-D object recognition is a challenging task that is at the core of many applications in robotics, indoor and outdoor. 
Nowadays, RGB-D sensors are ubiquitous in many robotic systems. They are inexpensive, widely supported by open source software, do not require complicated hardware and provide unique sensing capabilities. 
Compared to RGB data, which provides information about appearance and texture, depth data contains additional information about object shape and it is invariant to lighting or color variations.

In this paper, we propose a new method for object recognition from RGB-D data. In particular, we focus on making recognition robust to imperfect sensor data. A scenario typical for many robotics tasks.
\begin{figure}[tp]
  \centering
  \includegraphics[width=0.99\columnwidth]{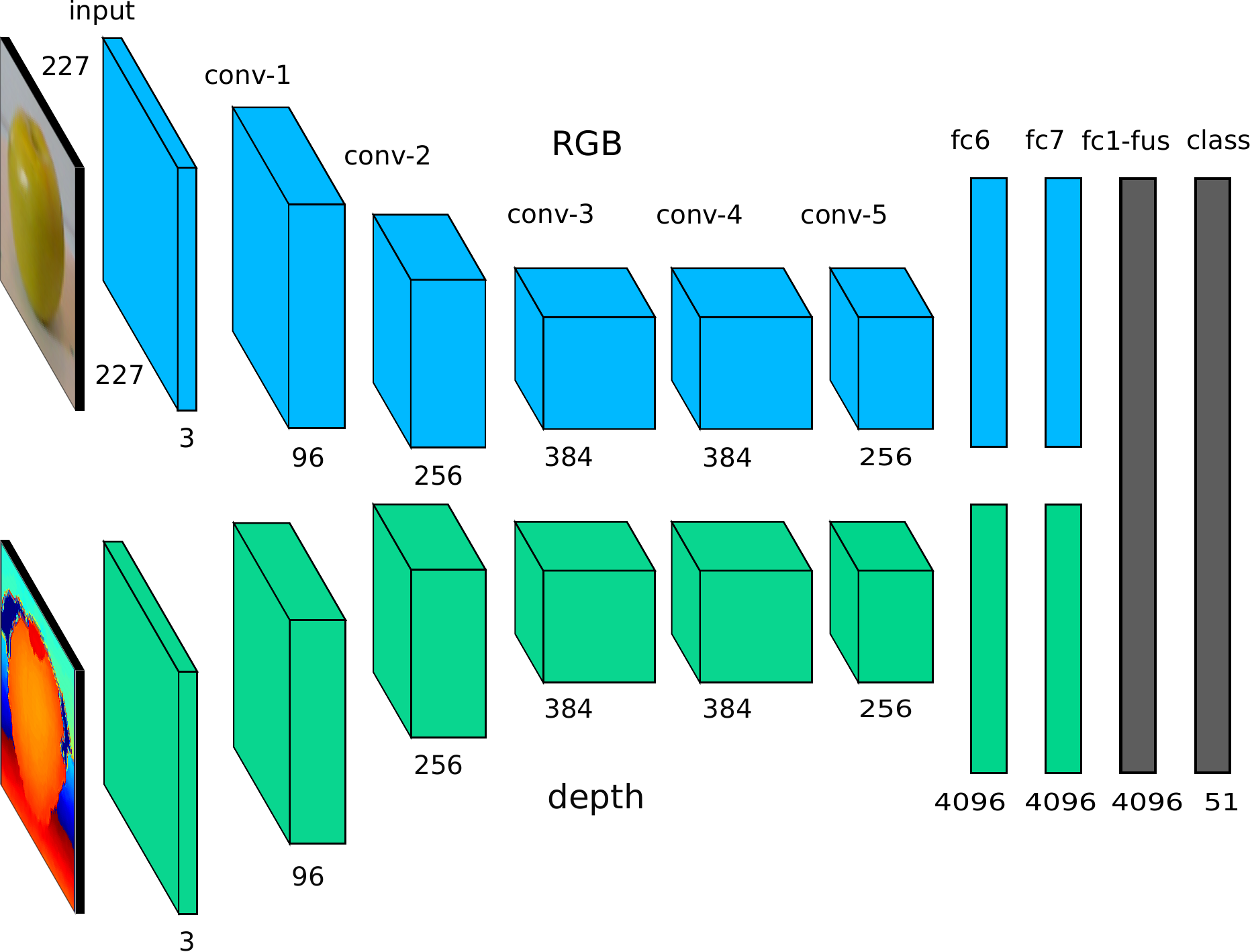}
  \caption{Two-stream convolutional neural network for RGB-D object recognition.  
    The input of the network is an RGB and depth image pair of size 227 $\times$ 227 $\times$ 3. Each stream (blue, green) consists of five convolutional layers and two fully connected layers. Both streams converge in one fully connected layer and a softmax classifier (gray).}
  \label{fig:aisnet-model}
\end{figure}
Our approach builds on recent advances from the machine learning and
computer vision community. Specifically, we extend classical convolutional
neural network networks (CNNs), which have recently been shown to be
remarkably successful for recognition on RGB images
\cite{Krizhevsky_imagenetclassification}, to the domain of RGB-D data.
Our architecture, which is depicted in
\figref{fig:aisnet-model}, consists of two convolutional network streams
operating on color and depth information respectively. The network
automatically learns to combine these two processing streams in a late
fusion approach.
This architecture bears similarity to other recent multi-stream
approaches
\cite{simonyan2014two,srivastava2012multimodal,BharathECCV2014}. 
Training
of the individual stream networks as well as the combined
architecture follows a stage-wise approach.
We start by separately training the networks for each modality, followed by a third training stage in which the two streams are 
jointly fine-tuned, together with a fusion network that performs the final classification.
We initialize both the RGB and depth stream network with weights from a network pre-trained on the ImageNet dataset~\cite{ILSVRCarxiv14}.
While initializing an RGB network from a pre-trained ImageNet network
is straight-forward, using such a network for processing depth data
is not. Ideally, one would want to directly train a network for
recognition from depth data without pre-training on a different modality which, however, is infeasible due to lack of large scale labeled depth datasets.
Due to this lack of labeled training data, a pre-training phase for the depth-modality -- leveraging RGB data -- becomes of key importance.
We therefore propose a depth data encoding to enable
re-use of CNNs trained on ImageNet for recognition from depth data.
The intuition -- proved experimentally -- is to simply encode a depth image
as a rendered RGB image, spreading the information contained in the
depth data over all three RGB channels and then using a standard
(pre-trained) CNN for recongition.

In real-world environments, objects are often subject to
occlusions and sensor noise. In this paper, we propose a data augmentation
technique for depth data that can be used for robust
training. We augment the available training examples by
corrupting the depth data with missing data patterns sampled from real-world environments.
Using these two techniques, our system can both learn robust depth features and implicitly weight the importance of the two modalities.

We tested our method to support our claims: first, we report on RGB-D recognition accuracy, then on robustness with respect to real-world noise.
For the first, we show that our work outperforms the current state of the art on the RGB-D Object
dataset of Lai \etal~\cite{lai2011large}.
For the second, we show that our data augmentation approach improves object
recognition accuracy in a challenging real-world and noisy 
environment using the RGB-D Scenes dataset~\cite{lai_icra12}.

  \begin{figure*}[t!]
    \centering
    \includegraphics[width=0.97\linewidth]{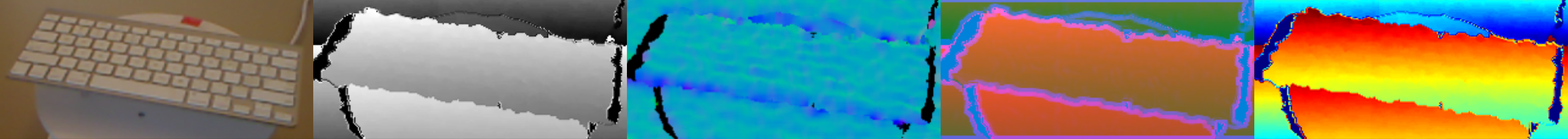}
    \caption{Different approaches for color encoding of depth images. From left to right: RGB, depth-gray,
    surface normals~\cite{Bo12unsupervised}, HHA~\cite{gupta2014learning}, our method.}
    \label{fig:depth-encode}
  \end{figure*}

\section{Related Work}
Our approach is related to a large body of work on both convolutional
neural networks (CNNs) for object recognition as well as applications
of computer vision techniques to the problem of recognition from RGB-D
data. Although a comprehensive review of the literature on CNNs and
object recognition is out of the scope of this paper, we will briefly
highlight connections and differences between our approach and
existing work with a focus on recent literature.

Among the many successful algorithms for RGB-D object
recognition a large portion still relies on hand designed features
such as SIFT in combination with multiple shape features on the depth
channel~\cite{lai2011large,bo2011depth}. However,
following their success in many computer vision problems, unsupervised
feature learning methods have recently been extended to RGB-D
recognition settings. Blum \etal~\cite{BlumSWR12} proposed an RGB-D
descriptor that relies on a K-Means based feature learning approach.
More recently Bo \etal~\cite{Bo12unsupervised} proposed hierarchical
matching pursuit (HMP), a hierarchical sparse-coding method that can
learn features from multiple channel input. A different approach
pursued by Socher \etal~\cite{SocherEtAl2012:CRNN} relies on combining
convolutional filters with a recursive neural network (a specialized
form of recurrent neural network) as the recognition architecture.
Asif \etal~\cite{asif2015icra} report improved recognition performance using a cascade of Random Forest classifiers that are fused in a hierarchical manner.
Finally, in recent independent work Schwarz \etal~\cite{SchwarzSB15} proposed to use features extracted from CNNs
pre-trained on ImageNet for RGB-D object recognition. While they
also make use of a two-stream network they do not fine-tune
the CNN for RGB-D recognition, but rather just use the pre-trained
network as is. Interestingly, they also discovered that simple
colorization methods for depth are competitive to more involved
preprocessing techniques. 
In contrast to their work, ours achieves higher accuracy by 
training our fusion CNN end-to-end: mapping from raw pixels to object classes in
a supervised manner (with pre-training on a related
recognition task). The features learned in our CNN are therefore by
construction discriminative for the task at hand.
Using CNNs trained for object recognition has a long history in
computer vision and machine learning. While they have been known to
yield good results on supervised image classification tasks such as
MNIST for a long time~\cite{Lecun98gradient-basedlearning}, recently they were
not only shown to outperform classical methods in large scale
image classification
tasks~\cite{Krizhevsky_imagenetclassification}, object
detection~\cite{girshick2014rcnn} and semantic
segmentation~\cite{10.1109/TPAMI.2012.231} but also to produce
features that transfer between
tasks~\cite{donahue2013decaf,AzizpourRSMC14}.
This recent success story
has been made possible through optimized implementations for
high-performance computing systems, as well as the availability of
large amounts of labeled image data through, e.g., the ImageNet
dataset~\cite{ILSVRCarxiv14}.

While the majority of work in deep learning has focused on 2D images,
recent research has also been directed towards using depth information
for improving scene labeling and object
detection~\cite{couprie-iclr-13, gupta2014learning}. Among them, the
work most similar to ours is the one on object detection by Gupta
\etal~\cite{gupta2014learning} who introduces a generalized method of
the R-CNN detector~\cite{girshick2014rcnn} that can be applied to
depth data. Specifically, they use large CNNs already
trained on RGB images to also extract features from depth data,
encoding depth information into three channels (HHA
encoding). Specifically, they encode for each pixel the height
  above ground, the horizontal disparity and the pixelwise angle between a
  surface normal and the gravity direction.
Our fusion network architecture shares similarities with their work in the 
usage of pre-trained networks on RGB images. Our method differs
in both the encoding of depth into color image data and in the fusion approach
taken to combine information from both modalities. For the encoding
step, we propose an encoding method for depth images ('colorizing'
depth) that does not rely on complicated preprocessing and results in
improved performance when compared to the HHA encoding. 
To accomplish sensor fusion we introduce additional layers to our CNN pipeline (see
\figref{fig:aisnet-model}) allowing us to automatically learn a
fusion strategy for the recognition task -- in contrast to simply
training a linear classifier on top of features extracted from both
modalities.
Multi-stream architectures have also been used for tasks such as action recognition~\cite{simonyan2014two}, detection~\cite{BharathECCV2014} and image retrieval~\cite{srivastava2012multimodal}.
An interesting recent overview of different network architectures for
fusing depth and image information is given in Saxena
\etal~\cite{lenz:deep}. There, the authors compared different
models for multimodal learning: (1) early fusion, in which the input
image is concatenated to the existing image RGB channels and processed
alongside; (2) an approach we denote as late fusion, where features
are trained separately for each modality and then merged at higher
layers; 
(3) combining early and late fusion;
concluding that late fusion (2) and the combined approach perform best for the problem of
grasp detection. 
Compared to their work, our model is similar to the late fusion approach but
widely differs in training -- Saxena \etal~\cite{lenz:deep} use a
layer-wise unsupervised training approach -- and scale (the size of
both their networks and input images is an order of magnitude smaller
than in our settings).

  \begin{figure}[t!]
    \centering
    \includegraphics[width=0.99\linewidth]{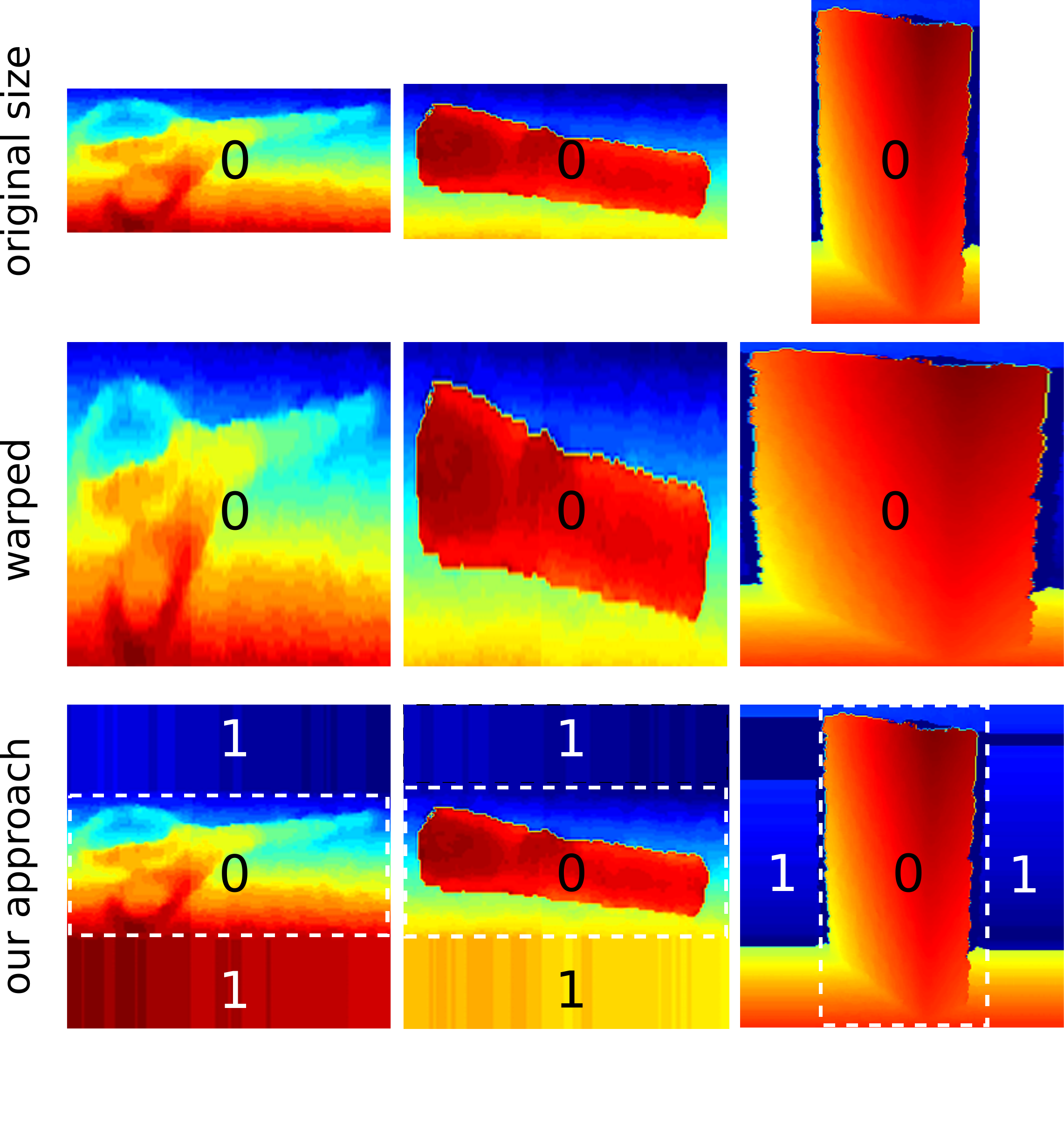}
    \caption{CNNs require a fixed size input. Instead of the widely
      used image warping approach (middle), our method (bottom)
      preserves shape information and ratio of the objects. We rescale the longer side and create additional
      image context, by tiling the pixels at the border of the longer
      side, e.g., 1.
      We assume that the depth image is already transformed to three
      channels using our colorization method.}
    \label{fig:borders_context}
  \end{figure}

\section{Multimodal architecture for RGB-D object recognition}
An overview of the architecture is given in
\figref{fig:aisnet-model}. Our network consists of two streams (top-blue
and bottom-green part in the figure) -- processing RGB and depth data
independently -- which are combined in a late fusion
approach. Each stream consists of a deep CNN that has been pre-trained for
object classification on the ImageNet database (we use the CaffeNet~\cite{jia2014caffe} implementation of the CNN from Krizhevsky \etal~\cite{Krizhevsky_imagenetclassification}).
The key reason behind starting from a pre-trained network is to enable
training a large CNN with millions of parameters using the limited
training data available from the Washington RGB-D Object dataset (see,
e.g., Yosinski \etal~\cite{Yosinski_2014} for a recent discussion).
We first pre-process data from both modalities to fully leverage the ImageNet pre-training.
Then, we train our multimodal CNN in a stage-wise manner. We fine-tune the parameters of each
individual stream network for classification of the target data and 
 proceed with the final training stage in which we jointly train the parameters of the
fusion network. 
The different steps will be outlined in the following sections.


\subsection{Input preprocessing}

To fully leverage the power of CNNs pre-trained on ImageNet, we pre-process the RGB and depth input data 
such that it is compatible with the kind of original ImageNet input.
Specifically, we use the reference implementation of the
CaffeNet~\cite{jia2014caffe} that expects 227 $\times$ 227 pixel RGB
images as input which are typically randomly cropped from larger 256
$\times$ 256 RGB images (see implementation details on data
augmentation). 
The first processing step consists of scaling the images to the appropriate image size.
The simplest approach to achieve this is to use image warping by directly
rescaling the original image to the required image dimensions, disregarding  
the original object ratio. 
This is depicted in 
\figref{fig:borders_context} (middle). We found in our experiments
that this process is detrimental to object recognition performance -- 
an effect that we attribute to a loss of shape information (see also Section \ref{sect:domain-adapt}). 
We therefore devise a different preprocessing approach: we scale the longest side of
the original image to 256 pixels, resulting in a 256 $\times$ N or an N
$\times$ 256 sized image. We then tile the borders of the longest side
along the axis of the shorter side. The resulting RGB or depth image
shows an artificial context around the object borders (see \figref{fig:borders_context}).
The same scaling operation is applied to both RGB and depth images.

While the RGB images can be directly used as inputs for the CNNs after
this processing step, the rescaled depth data requires additional steps.
To realize this, recall that a network trained on
ImageNet has been trained to recognize objects in images that follow a
specific input distribution (that of natural camera images) that is
incompatible with data coming from a depth sensor -- which essentially
encodes distance of objects from the sensor. Nonetheless, by looking at
a typical depth image from a household object scene (c.f.,
\figref{fig:scene}) one can conclude that many features that
qualitatively appear in RGB images -- such as edges, corners, shaded
regions -- are also visible in, e.g., a grayscale rendering of depth
data. This realization has previously led to the idea of simply using
a rendered version of the recorded depth data as an input for CNNs
trained on ImageNet~\cite{gupta2014learning}. We compare different
such encoding strategies for rendering depth to images in our
experiments. The two most prevalent such encodings are (1) 
rendering of depth data into grayscale and replicating the grayscale
values to the three channels required as network input; (2) using
surface normals where each dimension of a normal vector corresponds to
one channel in the resulting image. 
A more involved method, called HHA
encoding~\cite{gupta2014learning}, encodes in the three channels 
the height above ground, horizontal disparity and the
pixelwise angle between a surface normal and the gravity direction.

We propose a fourth, effective and computationally inexpensive, encoding of depth to color images, which we
found to outperform the HHA encoding for object recognition.
Our method first normalizes all depth values to lie between 0 and 255.
Then, we apply a jet colormap on the given image that transforms the
input from a single to a three channel image (colorizing the depth).
For each pixel $(i,j)$ in the depth image $d$ of
size $W \times H$, we map the distance to color values ranging from red
(near) over green to blue (far), essentially distributing the depth
information over all three RGB channels. Edges in these three channels
often correspond to interesting object boundaries. Since the network is
designed for RGB images, the colorization procedure provides enough
common structure between a depth and an RGB image to learn suitable
feature representations (see \figref{fig:depth-encode} for
a comparison between different depth preprocessing methods).

\subsection{Network training}
Let $\mathcal{D} = \lbrace (\bx^{1}, \bd^{1}, \by^{1}), \dots,
(\bx^{N}, \bd^{N}, \by^{N}) \rbrace$ be the labeled data available for
training our multimodal CNN; with $\bx^{i},\bd^{i}$ denoting the RGB
and pre-processed depth image respectively and $\by^{i}$ corresponding
to the image label in one-hot encoding -- i.e., $\by^{i} \in
\mathbb{R}^M$ is a vector of dimensionality M (the number of labels)
with $y^i_k = 1$ for the position $k$ denoting the image label. We
train our model using a three-stage approach, first training the two
stream networks individually followed by a joint fine-tuning stage.

\subsubsection{Training the stream networks}
\label{sect:train-individual} We first proceed by training the two
individual stream networks (c.f., the blue and green streams in
\figref{fig:aisnet-model}). Let $g^{I}(\bx^{i}; \theta^I)$ be the
representation extracted from the last fully connected layer (fc7) of
the CaffeNet -- with parameters $\theta^I$ -- when applied to an RGB
image $\bx^i$. Analogously, let $g^{D}(\bd^{i}; \theta^D)$ be the
representation for the depth image. We will assume that all parameters
$\theta^I$ and $\theta^D$ (the network weights and biases) are
initialized by copying the parameters of a CaffeNet trained on the
ImageNet dataset. We can then train an individual stream network by
placing a randomly initialized softmax classification layer on top of
$f^D$ and $f^I$ and minimizing the negative log likelihood
$\mathcal{L}$ of the training data. That is, for the depth image stream
network we solve \begin{equation}
  \min_{\bW^D, \theta^D} \sum_{i=1}^{N} \mathcal{L}\left (\text{softmax} \left (\bW^D g^{D}(\bd^{i}; \theta^D) \right), y^i \right),
\label{eq:objective-stream}
\end{equation}
where $W^D$ are the weights of the softmax layer mapping from
$g(\cdot)$ to $\mathbb{R}^M$, the softmax function is given by
$\text{softmax}(\bz) = \text{exp}(\bz) / \| \bz \|_1$ and the loss is
computed as $\mathcal{L}(s, y) = - \sum\limits_{k} y_k \log s_k$.
Training the RGB stream network then can be performed by an analogous
optimization. After training, the resulting networks can be used to
perform separate classification of each modality.

\subsubsection{Training the fusion network} Once the two individual
stream networks are trained we discard their softmax weights,
concatenate their -- now fine-tuned -- last layer responses
$g^{I}(\bx^{i}; \theta^I)$ and $g^{D}(\bd^{i}; \theta^D)$ and feed
them through an additional fusion stream $f([ g^{I}(\bx^{i};
\theta^I), g^{D}(\bd^{i}; \theta^D)]; \theta^F)$ with parameters
$\theta^F$. This fusion network again ends in a softmax classification
layer. The complete setup is depicted in 
\figref{fig:aisnet-model}, where the two fc7 layers (blue and green) are
concatenated and merge into the fusion network (here the inner product layer fc1-fus depicted in
gray). Analogous to \eqref{eq:objective-stream} the fusion network can
therefore be trained by jointly optimizing all parameters to minimize
the negative log likelihood \begin{equation}
  \min_{\substack \bW^f, \theta^I, \theta^D, \theta^F} \sum_{i=1}^{N} \mathcal{L}\left (\text{softmax} \left (\bW^f f([\bg^{I},  \bg^{D}]; \theta^F) \right), y^i \right),
\label{eq:objective-fusion}
\end{equation}
where $\bg^{I} = g^{I}(\bx^{i}; \theta^I)$, $\bg^{D} = g^{D}(\bd^{i};
\theta^D)$. Note that in this stage training can also be performed by
optimizing only the weights of the fusion network (effectively keeping
the weights from the individual stream training intact).

  \begin{figure}[tb]
    \centering
    \includegraphics[width=0.90\linewidth]{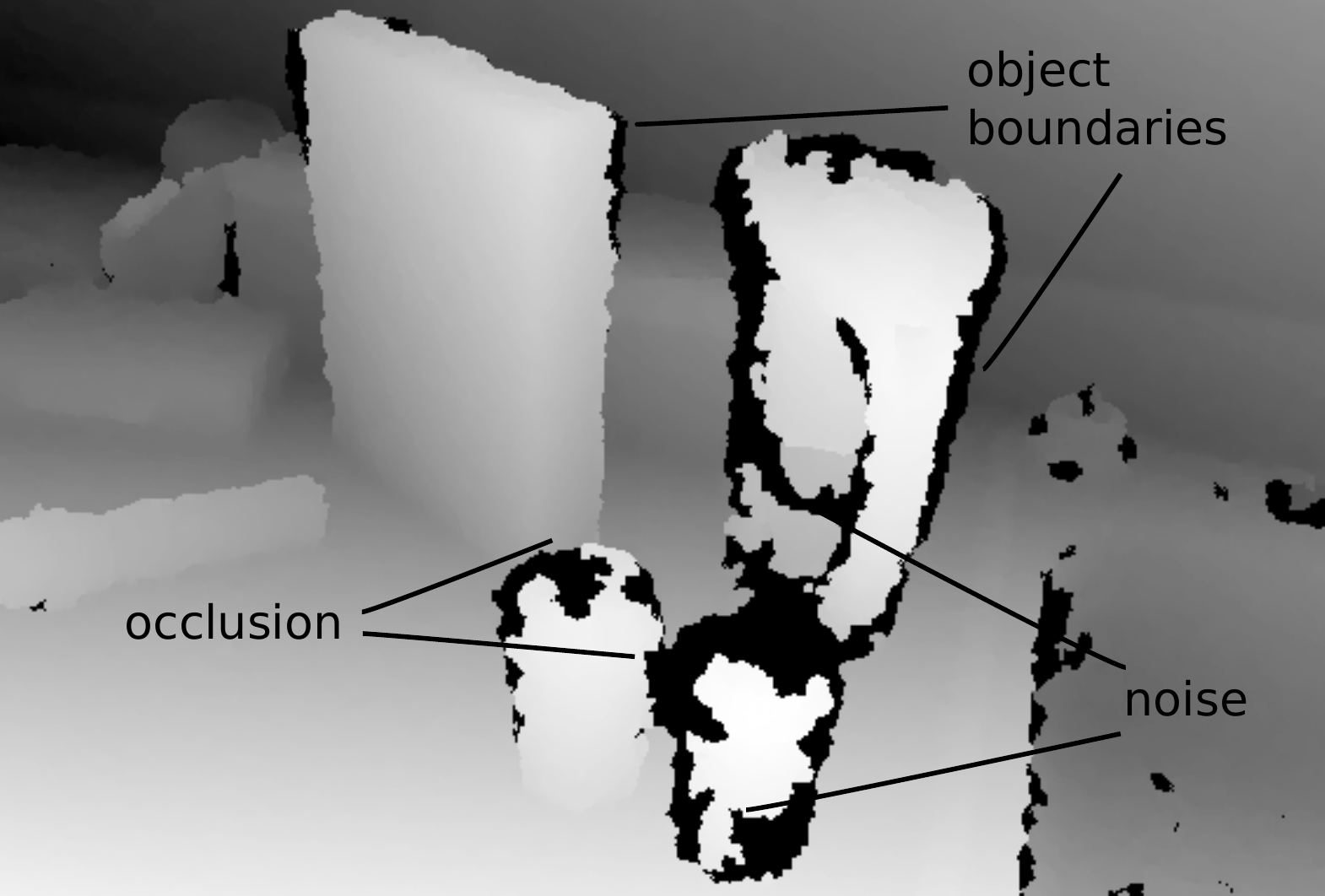}
    \caption{Kitchen scene in the RGB-D Scenes dataset showing objects subjected to noise and occlusions.}
    \label{fig:scene}
  \end{figure}

  \begin{figure}
    \centering
    \includegraphics[width=0.90\columnwidth]{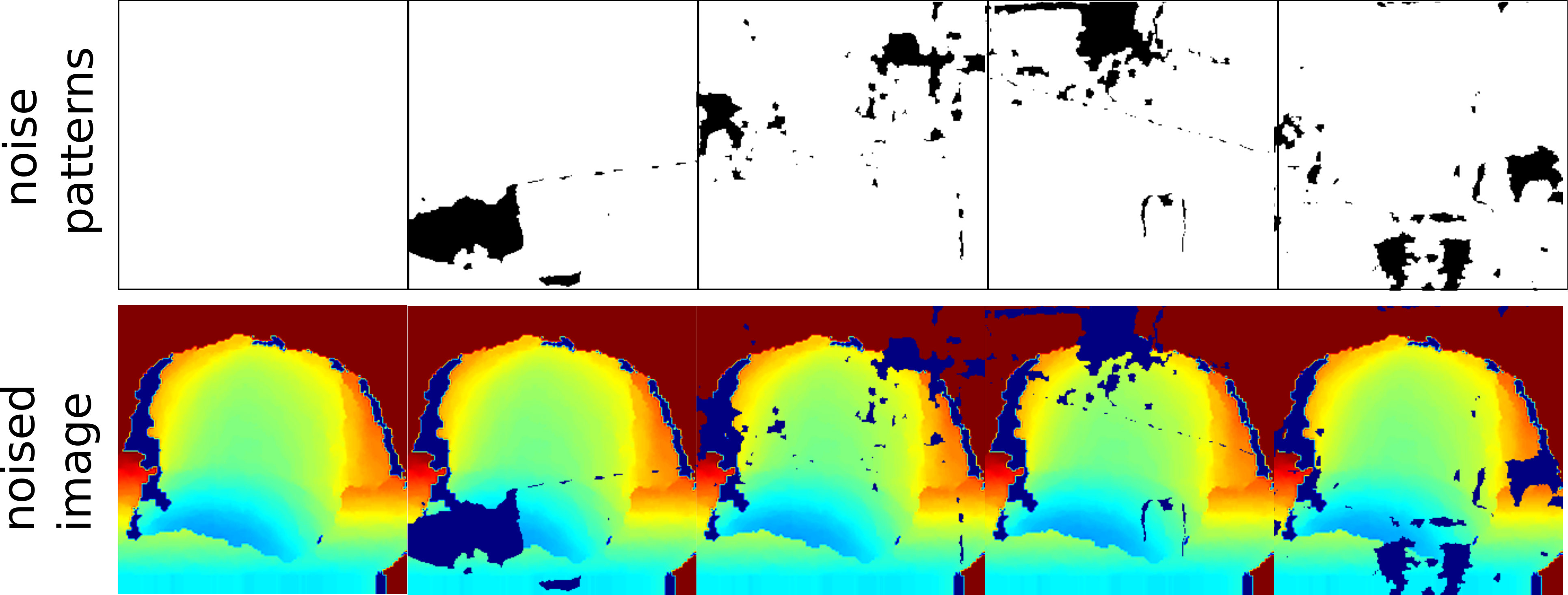}
    \caption{We create synthetic training data by inducing artificial
      patterns of missing depth information in the encoded
      image.}
    \label{fig:noise-samples}
  \end{figure}

\subsection{Robust classification from depth images}
\label{sect:train-robust}
Finally, we are interested in using our approach in real world
robotics scenarios. Robots are supposed to perform object recognition in cluttered scenes where 
the perceived sensor data is
subject to changing external conditions (such as lighting) and sensor
noise. 
Depth sensors are especially affected by a 
non-negligible amount of noise in such setups. This is mainly due to the fact that reflective properties 
of materials as well as their coating, often result in missing depth information.
An example of noisy depth data is depicted in \figref{fig:scene}. In
contrast to the relatively clean training data from the Washington RGB-D
Object dataset, the depicted scene contains considerable amounts of
missing depth values and partial occlusions (the black pixels in
the figure). To achieve robustness against such unpredictable factors, 
we propose a new data augmentation scheme that generates new, noised
training examples for training and is tailored specifically to robust
classification from depth data.

Our approach utilizes the observation that noise in depth data often
shows a characteristic pattern and appears at object boundaries or
object surfaces. Concretely, we sampled a representative set of noise
patterns $\mathcal{P} = \lbrace P_1, \dots, P_K \rbrace$ that occur
when recording typical indoor scenes through a Kinect sensor. For
sampling the noise patterns we used the RGB-D SLAM
dataset~\cite{sturm12iros}. First, we extract 33,000 random noise
patches of size 256 $\times$ 256 from different sequences at varying
positions and divide them into five groups, based on the number of
missing depth readings they contain. Those noise patches are 2D binary masks patterns.
We randomly sample pairs of
noise patches from two different groups that are randomly added 
or subtracted and optionally inverted to produce a final noise
mask pattern. We repeat this process until we have collected $K = 50,000$
noise patterns in total. Examples of the resulting noise patterns and
their application to training examples are shown in
~\figref{fig:noise-samples}.

Training the depth network with artificial noise patterns then
proceeds by minimizing the objective from Equation
\eqref{eq:objective-stream} in which each depth sample $\bd^i$ is randomly replaced
with a noised variant with probability $50\%$. Formally,
\begin{equation}
  \bd^i = \begin{cases} 
  \bd^i &\text{ if } p = 1   \\
  P_k \circ \bd^i &\text{ else }
\end{cases} \text{ with } \begin{aligned} 
  p &\sim \mathcal{B}\lbrace 0.5 \rbrace \\
  k &\sim \mathcal{U}\lbrace 1,K \rbrace,
\end{aligned}
\end{equation}
where $\circ$ denotes the Hadamard product, $\mathcal{B}$
the Bernoulli distribution and $\mathcal{U}$ the discrete uniform distribution.

\section{Experiments}
We evaluate our multimodal network architecture on the Washington
RGB-D Object Dataset~\cite{lai2011large} which consists of household
objects belonging to 51 different classes. As an additional experiment
-- to evaluate the robustness of our approach for classification in
real-world environments -- we considered classification of objects
from the RGB-D Scenes dataset whose class distribution partially
overlaps with the RGB-D Object Dataset.

\subsection{Experimental setup}
All experiments were performed using the publicly available Caffe
framework~\cite{jia2014caffe}. As described previously we use the CaffeNet as the basis for our fusion network. 
It consists of five convolutional layers (with max-pooling after the first, second and
fifth convolution layer) followed by two fully connected layers and a
softmax classification layer. Rectified linear units are used in all
but the final classification layer. We initialized both stream
networks with the weights and biases of the first eight layers from this
pre-trained network, discarding the softmax layer. We then proceeded
with our stage-wise training. In the first stage (training the RGB and
depth streams independently) the parameters of all layers were adapted
using a fixed learning rate schedule (with initial learning rate of
0.01 that is reduced to 0.001 after 20K iterations and training is stopped after 30K iterations). 
In the second stage (training the fusion network, 20k iterations, mini-batch size of 50) we experimented with fine-tuning all weights but found that fixing the individual stream networks (by
setting their learning rate to zero) and only training the fusion part of
the network resulted in the best performance.
The number of training iterations were chosen based on the validation performance on a training validation split in a preliminary experiment.
A fixed momentum value of 0.9 and a mini-batch size of 128 was used for all experiments if not stated otherwise. We
also adopted the common data augmentation practices of randomly
cropping 227 $\times$ 227 sub-images from the larger 256 $\times$ 256
input examples and perform random horizontal flipping. Training of a
single network stream takes ten hours, using a NVIDIA 780 graphics
card.

\subsection{RGB-D Object dataset}

  \begin{table}[t!]
    \centering
    \caption{Comparisons of our fusion network with other approaches reported for the RGB-D dataset. Results are recognition accuracy in percent. Our multi-modal CNN outperforms all the previous approaches.}
    \label{tab:x-comparison}
    \begin{tabular}{c|c|c|c}
      \hline
      Method & RGB & Depth & RGB-D \\
      \hline
      Nonlinear SVM~\cite{lai2011large} & 74.5 $\pm$ 3.1 & 64.7 $\pm$ 2.2 & 83.9 $\pm$ 3.5  \\
      HKDES ~\cite{bo2011object} & 76.1 $\pm$ 2.2 & 75.7 $\pm$ 2.6 & 84.1 $\pm$ 2.2  \\
      Kernel Desc.~\cite{bo2011depth} & 77.7 $\pm$ 1.9 & 78.8 $\pm$ 2.7 & 86.2 $\pm$ 2.1  \\
      CKM Desc.~\cite{BlumSWR12}  & N/A & N/A & 86.4 $\pm$ 2.3 \\
      CNN-RNN~\cite{SocherEtAl2012:CRNN}  & 80.8 $\pm$ 4.2 & 78.9 $\pm$ 3.8 & 86.8 $\pm$ 3.3 \\
      Upgraded HMP~\cite{Bo12unsupervised}  & 82.4 $\pm$ 3.1  & 81.2  $\pm$ 2.3  & 87.5 $\pm$ 2.9\\
      
      CaRFs~\cite{asif2015icra} & N/A & N/A & 88.1 $\pm$ 2.4 \\

      CNN Features~\cite{SchwarzSB15} & 83.1 $\pm$ 2.0 & N/A & 89.4 $\pm$ 1.3 \\ 

      \hline
      \hline
      Ours, Fus-CNN (HHA) & {\bf84.1 $\pm$ 2.7} & {\bf83.0 $\pm$ 2.7} & {\bf91.0 $\pm$ 1.9} \\
      Ours, Fus-CNN (jet) & {\bf84.1 $\pm$ 2.7} & {\bf83.8 $\pm$ 2.7} & {\bf91.3 $\pm$ 1.4} \\
    \end{tabular}
\end{table}

The Washington RGB-D Object Dataset consists of 41,877 RGB-D images
containing household objects organized into 51 different classes and a
total of 300 instances of these classes which are captured under three
different viewpoint angles. For the evaluation every 5th frame
is subsampled. We evaluate our method on the challenging category
recognition task, using the same ten cross-validation splits as in
Lai \etal~\cite{lai2011large}. Each split consists of roughly 35,000
training images and 7,000 images for testing. From each object class
one instance is left out for testing and training is performed on
the remaining $300-51=249$ instances. At test time the task of the
CNN is to assign the correct class label to a previously unseen
object instance.

\begin{figure}[t!]
    \centering
    \includegraphics[width=0.99\linewidth]{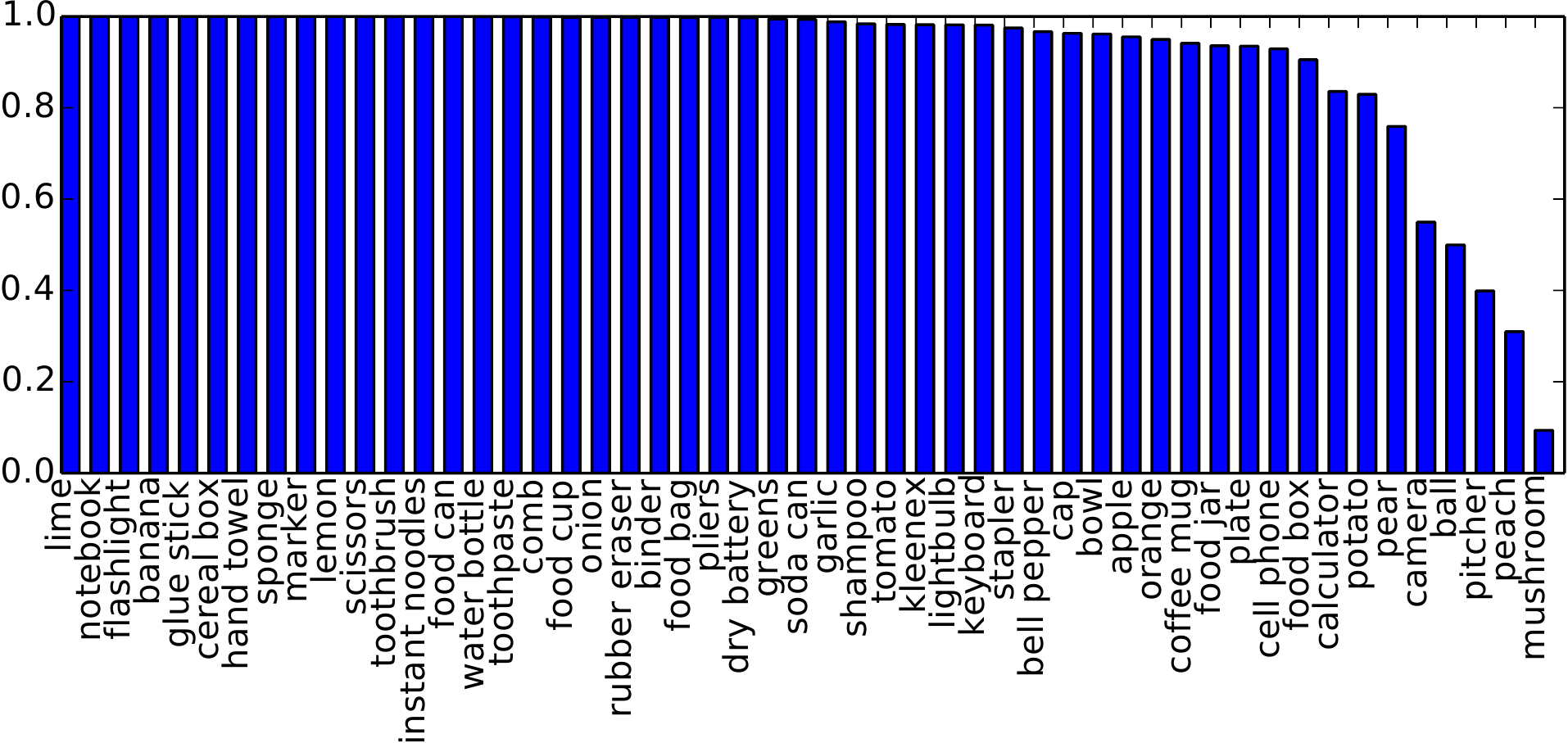}
    \caption{Per-class recall of our trained model on all test-splits. The worst class recall belongs to mushrooms and peaches.}
    \label{fig:classrecall}
  \end{figure}

Table~\ref{tab:x-comparison} shows the average accuracy of our
multi-modal CNN in comparison to the best results reported in the
literature. 
Our best multi-modal CNN, using the jet-colorization, (Fus-CNN jet) yields
an overall accuracy of $91.3 \pm 1.4\%$ when using RGB and depth
($84.1 \pm 2.7\%$ and $83.8 \pm 2.7\%$ when only the RGB or depth
modality is used respectively), which -- to the best of our knowledge
-- is the highest accuracy reported for this dataset to date. We also
report results for combining the more computationally intensive HHA with our network 
(Fus-CNN HHA). As can be seen in the table, this
did not result in an increased performance. The depth
colorization method slightly outperforms the HHA fusion network (Fus-CNN
HHA) while being computationally cheaper.
Overall our experiments show that a pre-trained CNN can be adapted for
recognition from depth data using our depth colorization method.
Apart from the results reported in the table, we also experimented with
different fusion architectures. Specifically, performance slightly
drops to $91\%$ when the intermediate fusion layer (fc1-fus) is
removed from the network. Adding additional fusion layers also did not
yield an improvement. 
Finally, 
\figref{fig:classrecall} shows the per-class recall, where roughly
half of the objects achieve a recall of $\approx 99\%$.

 
\subsection{Depth domain adaptation for RGB-D Scenes}
\label{sect:domain-adapt}
To test the effectiveness of our depth
augmentation technique in real
world scenes, we performed additional recognition experiments on the more
challenging RGB-D Scenes dataset. This dataset consists of six object
classes (which overlap with the RGB-D Object Dataset) and a large
amount of depth images subjected to noise.

\begin{figure}[t!]
    \centering
        \includegraphics[width=0.49\textwidth]{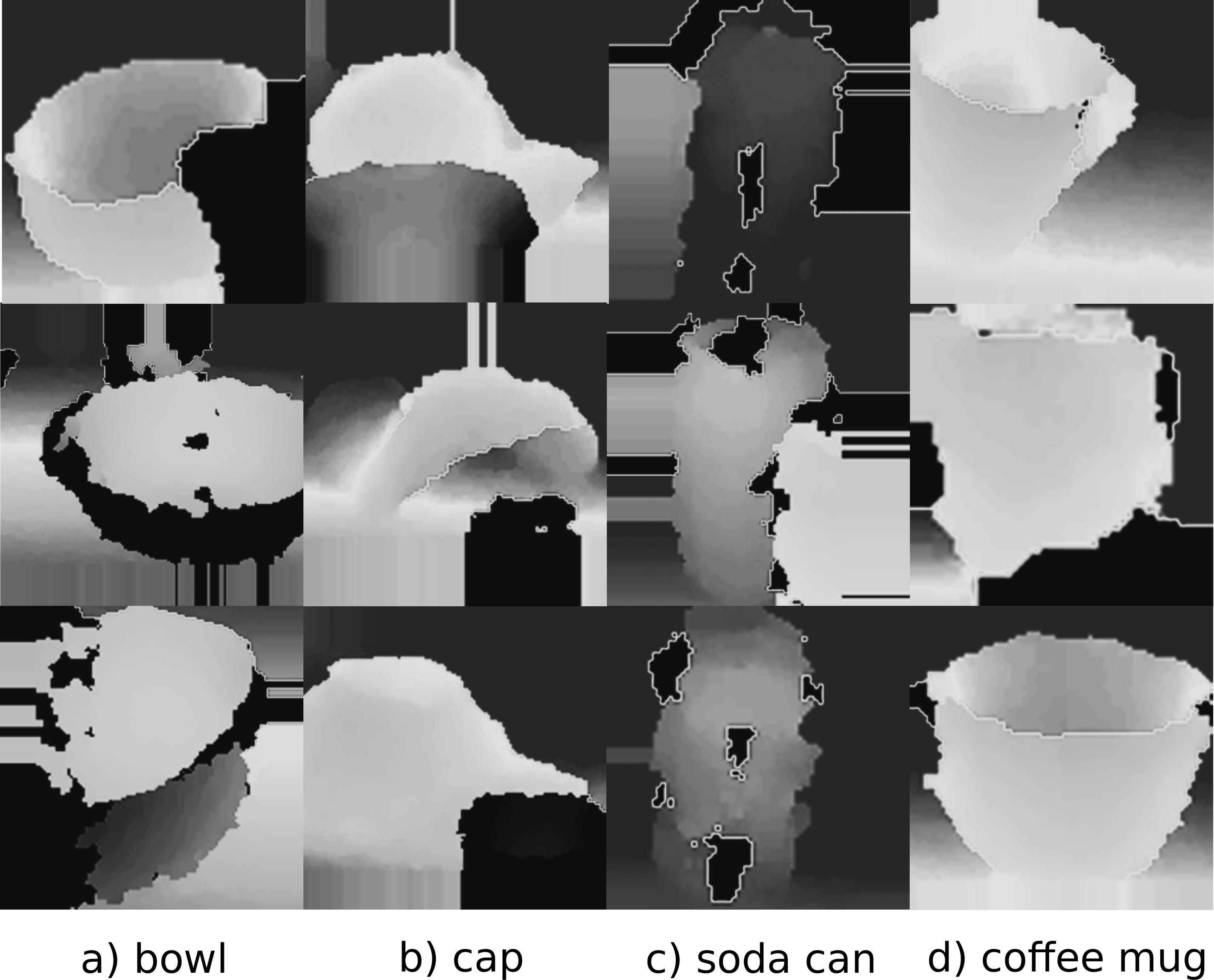}
        \label{fig:bowl}
        \caption{Objects from the RGB-D Scenes test-set for
          which the domain adapted CNN predicts the correct label,
          while the baseline (no adapt.) CNN fails. Most of these
          examples are subject to noise or partial occlusion.}
    \label{fig:soda_can}
  \end{figure}

For this experiment we trained two single-stream depth-only networks using the Object dataset and used the Scenes dataset for testing.
Further, we assume that the groundtruth bounding box is given in order to report only on recognition performance.
The first ``baseline'' network is trained by following the
procedure described in Section \ref{sect:train-individual}, with the
total number of labels $M=6$. The second network is trained by
making use of the depth augmentation outlined in
\ref{sect:train-robust}. The results of this experiment are shown in
Table~\ref{tab:scenes-comparison} (middle and right column) that reports
the recognition accuracy for each object class averaged over all eight
video sequences. As is evident from the table, the adapted network
(right column) trained with data augmentation outperforms the baseline
model for all classes, clearly indicating that additional domain
adaptation is necessary for robust recognition in real world scenes.
However, some classes (e.g., cap, bowl, soda can) benefit more from
noise aware training than others (e.g., flashlight, coffe mug). The
kitchen scene depicted in \figref{fig:scene} gives a visual intuition
for this result. On the one hand, some objects (e.g., soda cans) often
present very noisy object boundaries and surfaces, thus they show 
improved recognition performance using the adapted approach. 
On the
other hand, small objects (e.g. a flashlight), which are often captured
lying on a table, are either less noisy or just small, hence susceptible to be completely erased by the noise from
our data augmentation approach.
\figref{fig:soda_can} shows several
exemplary noisy depth images from the test set that are correctly classified by the domain-adapted
network while the baseline network labels them incorrectly.
We also tested the effect of different input image rescaling
techniques -- previously described in \figref{fig:borders_context} --
in this setting. As shown in the left column of
Table~\ref{tab:scenes-comparison}, standard image warping performs
poorly, which supports our intuition that shape information gets lost
during preprocessing.

  \begin{table}[t]
    \centering
    \caption{Comparison of the domain adapted depth network with the baseline: six-class recognition results (in percent) on the RGB-D Scenes dataset~\cite{lai_icra12} that contains everyday objects in real-world environments.}
    \label{tab:scenes-comparison}
    \begin{tabular}{c|c|c|c}
      \hline
      Class  & Ours, warp. & Ours, no adapt. & Ours, adapt.\\      
      \hline 
      flashlight   &   93.4   & \bf97.5  &  96.4  \\
      cap          &   62.1  & 68.5  & \bf77.4 \\
      bowl         &   57.4  & 66.5  & \bf69.8  \\ 
      soda can     &   64.5 & 66.6  & \bf71.8 \\
      cereal box   &   \bf98.3  & 96.2  & 97.6 \\
      coffee mug   &   61.9  & 79.1  & \bf79.8 \\
      \hline
      \textbf{class avg.}   & 73.6 $\pm$ 17.9  &  79.1 $\pm$ 14.5  & \bf82.1 $\pm$ 12.0 \\
    \end{tabular}
  \end{table}

\subsection{Comparison of depth encoding methods}
Finally, we conducted experiments to compare the different
depth encoding methods described in \figref{fig:depth-encode}. 
For rescaling the images, we use our
proposed preprocessing method described
in~\figref{fig:borders_context} and tested the different depth encoding. 
Two scenarios are considered: 1) training from scratch using single channel depth images 2) for each encoding method, only fine-tuning the network by using the procedure described in 
Section \ref{sect:train-individual}.
When training from scratch, the initial learning rate is set to 0.01, then changed to 0.001 after 40K iterations thus stopped after 60K iterations. Training with more iterations did not further improve the accuracy.
From the results, presented in Table~\ref{tab:encode-comparison}, it is clear that training the network from scratch -- solely on the RGB-D Dataset -- is inferior to fine-tuning.    
In the latter setting, the results suggest that the simplest encoding method
(depth-gray) performs considerably worse than the other three methods.
Among these other encodings (which all produce colorized images), 
surface normals and HHA encoding require additional image
preprocessing -- meanwhile colorizing depth using our depth-jet encoding has negligible computational overhead.
One potential reason why the HHA encoding underperforms in this setup is that all objects are
captured on a turntable with the same height above the ground. The
height channel used in the HHA encoding therefore does not encode any
additional information for solving the classification task.
In this experiment, using surface normals yields slightly better performance than the depth-jet encoding.
Therefore, we tested the fusion architecture on the ten splits of the RGB-D Object Dataset using the surface normals encoding but this did not further improve the performance. Specifically, the recognition accuracy on the test-set was 91.1 $\pm$ 1.6 which is comparable to our reported results in Table~\ref{tab:x-comparison}.

\section{Conclusion}

We introduce a novel multimodal neural network archi-
tecture for RGB-D object recognition, which achieves state
of the art performance on the RGB-D Object dataset \cite{lai2011large}.
Our method consists of a two-stream convolutional neural
network that can learn to fuse information from both RGB
and depth automatically before classification. We make use
of an effective encoding method from depth to image data
that allows us to leverage large CNNs trained for object
recognition on the ImageNet dataset. We present a novel
depth data augmentation that aims at improving recognition
in noisy real-world setups, situations typical of many robotics
scenarios. We present extensive experimental results and confirm that our method
is accurate and it is able to learn rich features from both
domains. We also show robust object recognition in real-
world environments and prove that noise-aware training is effective
and improves recognition accuracy on the RGB-D Scenes
dataset~\cite{lai_icra12}.
  \begin{table}[t!]
    \centering
    \caption{Comparison of different depth encoding methods on the ten test-splits of the RGB-D Object dataset.}
    \label{tab:encode-comparison}
    \begin{tabular}{c|c}
      \hline
      Depth Encoding       &     Accuracy   \\
      \hline
      Depth-gray (single channel), from scratch  &  80.1 $\pm$ 2.6 \\   
      Depth-gray          &     82.0 $\pm$ 2.8     \\
      Surface normals         &     \bf{84.7} $\pm$ 2.3    \\
      HHA             &     83.0 $\pm$ 2.7    \\
      Depth-jet encoding           &     83.8 $\pm$ 2.7    \\
    \end{tabular}
  \end{table}




\IEEEtriggeratref{4}
\IEEEtriggercmd{\enlargethispage{-0.2in}}
\bibliographystyle{IEEEtrans}
\bibliography{references}
\addtolength{\textheight}{-12cm}   

\end{document}